\documentclass{article}
\usepackage[nonatbib,final]{neurips_data_2022}

\usepackage[utf8]{inputenc} %
\usepackage[T1]{fontenc}    %
\usepackage[numbers]{natbib}

\usepackage{amsmath,amsfonts,bm}

\def\eqref#1{equation~\ref{#1}}

\def\1{\bm{1}}

\DeclareMathAlphabet{\mathsfit}{\encodingdefault}{\sfdefault}{m}{sl}
\SetMathAlphabet{\mathsfit}{bold}{\encodingdefault}{\sfdefault}{bx}{n}

\def\gE{{\mathcal{E}}}

\def\gG{{\mathcal{G}}}

\def\gV{{\mathcal{V}}}

\usepackage{color,xcolor}
\usepackage{epsfig}
\usepackage{graphicx}

\usepackage{adjustbox}
\usepackage{array}
\usepackage{booktabs}
\usepackage{colortbl}
\usepackage{wrapfig}
\usepackage{hhline}
\usepackage{multirow}
\usepackage{subcaption} %
\usepackage{wrapfig}
\usepackage{floatflt}
\usepackage{pdfpages}
\usepackage{amsmath,amsfonts,amssymb}
\usepackage{mathtools}  %
\usepackage{bm}
\usepackage{nicefrac}
\usepackage{microtype}

\usepackage{changepage}
\usepackage{extramarks}
\usepackage{fancyhdr}
\usepackage{lastpage}
\usepackage{setspace}
\usepackage{soul}
\usepackage{xspace}

\usepackage[pagebackref=true,breaklinks=true,colorlinks,citecolor=gray]{hyperref}

\usepackage{url}

\usepackage{enumerate}
\usepackage{todonotes} %
\usepackage{enumitem}  %

\usepackage{titlesec}

\usepackage{makecell}

\usepackage{pifont} %

\usepackage{algorithm,algpseudocode}

\usepackage{amsthm}
\usepackage{framed}
\definecolor{shadecolor}{gray}{0.95}
\newcolumntype{L}[1]{>{\raggedright\let\newline\\\arraybackslash\hspace{0pt}}m{#1}}
\newcolumntype{C}[1]{>{\centering\let\newline\\\arraybackslash\hspace{0pt}}m{#1}}
\newcolumntype{R}[1]{>{\raggedleft\let\newline\\\arraybackslash\hspace{0pt}}m{#1}}

\newcommand{\sect}[1]{Section~\ref{#1}}

\newcommand{\fig}[1]{Fig.~\ref{#1}}
\newcommand{\tbl}[1]{Table~\ref{#1}}

\newcommand{\ignore}[1]{}

\makeatletter
\DeclareRobustCommand\onedot{\futurelet\@let@token\@onedot}
\def\@onedot{\ifx\@let@token.\else.\null\fi\xspace}

\def\eg{e.g\onedot} 
\def\ie{i.e\onedot} 
 
\def\etc{etc\onedot}

\makeatother

\definecolor{MyDarkBlue}{rgb}{0,0.08,1}
\definecolor{MyDarkGreen}{rgb}{0.02,0.6,0.02}
\definecolor{MyDarkRed}{rgb}{0.8,0.02,0.02}
\definecolor{MyDarkOrange}{rgb}{0.40,0.2,0.02}
\definecolor{MyPurple}{RGB}{111,0,255}
\definecolor{MyRed}{rgb}{1.0,0.0,0.0}
\definecolor{MyGold}{rgb}{0.75,0.6,0.12}
\definecolor{MyDarkgray}{rgb}{0.66, 0.66, 0.66}

\newcommand{\dataset}{CLEVRER-Humans\xspace}

\newcommand{\xhdr}[1]{\noindent\textbf{#1}}

\newcommand{\mycellc}[1]{\begin{tabular}[t]{@{}c@{}l}#1\end{tabular}}

\usepackage[normalem]{ulem}

\title{CLEVRER-Humans: Describing Physical and\\ Causal Events the Human Way}
\author{%
  Jiayuan Mao\thanks{indicates equal contribution. Authors ordered alphabetically. Correspondence to: {\tt jiayuanm@mit.edu}.\\
  Project page: \url{https://sites.google.com/stanford.edu/clevrer-humans/home}.\\Dataset DOI: \url{https://doi.org/10.5061/dryad.5tb2rbp7c}.} \\
  MIT\\
  \And 
  Xuelin Yang$^*$ \\ 
  Stanford University\\
\And
  Xikun Zhang \\
  Stanford University\\
  \AND
  Noah D. Goodman \\
  Stanford University\\
  \And
  Jiajun Wu \\
  Stanford University\\
}

\begin{document}

\theoremstyle{plain}
\newtheorem{theorem}{Theorem}[section]
\newtheorem{proposition}[theorem]{Proposition}
\newtheorem{lemma}[theorem]{Lemma}
\newtheorem{corollary}[theorem]{Corollary}
\theoremstyle{definition}
\newtheorem{definition}[theorem]{Definition}
\newtheorem{assumption}[theorem]{Assumption}
\theoremstyle{remark}
\newtheorem{remark}[theorem]{Remark}

\maketitle

\begin{abstract}
Building machines that can reason about physical events and their causal relationships is crucial for flexible interaction with the physical world. However, most existing physical and causal reasoning benchmarks are exclusively based on synthetically generated events and synthetic natural language descriptions of causal relationships. This design brings up two issues. First, there is a lack of diversity in both event types and natural language descriptions; second, causal relationships based on manually-defined heuristics are different from human judgments. To address both shortcomings, we present the CLEVRER-Humans benchmark, a video reasoning dataset for causal judgment of physical events with human labels. We employ two techniques to improve data collection efficiency: first, a novel iterative event cloze task to elicit a new representation of events in videos, which we term Causal Event Graphs (CEGs); second, a data augmentation technique based on neural language generative models. We convert the collected CEGs into questions and answers to be consistent with prior work. Finally, we study a collection of baseline approaches for CLEVRER-Humans question-answering, highlighting the great challenges set forth by our benchmark\footnote{The latest dataset version is version 3, which has been updated on May 12, 2023.}.
\end{abstract}

\section{Introduction}
The ability to reason about physical events and their causal relationships from visual observations lies at the core of human intelligence. It is crucial for humans to holistically understand and flexibly interact with the physical world \citep{talmy1988force,wolff2007representing,gerstenberg2012noisy,battaglia2013simulation,gerstenberg2017intuitive}.
Natural language provides a way for humans to express such causal understanding \citep{beller2020language}, and we can use natural language as a lens to evaluate machine understanding of physical events and causal judgments. This brings us two major advantages. First, compared to bounding boxes and timecodes, language provides a more flexible interface for describing events. Furthermore, it naturally enables the generation of human-interpretable explanations.

\begin{figure}[!tp]
    \centering\small
    \includegraphics[width=\textwidth]{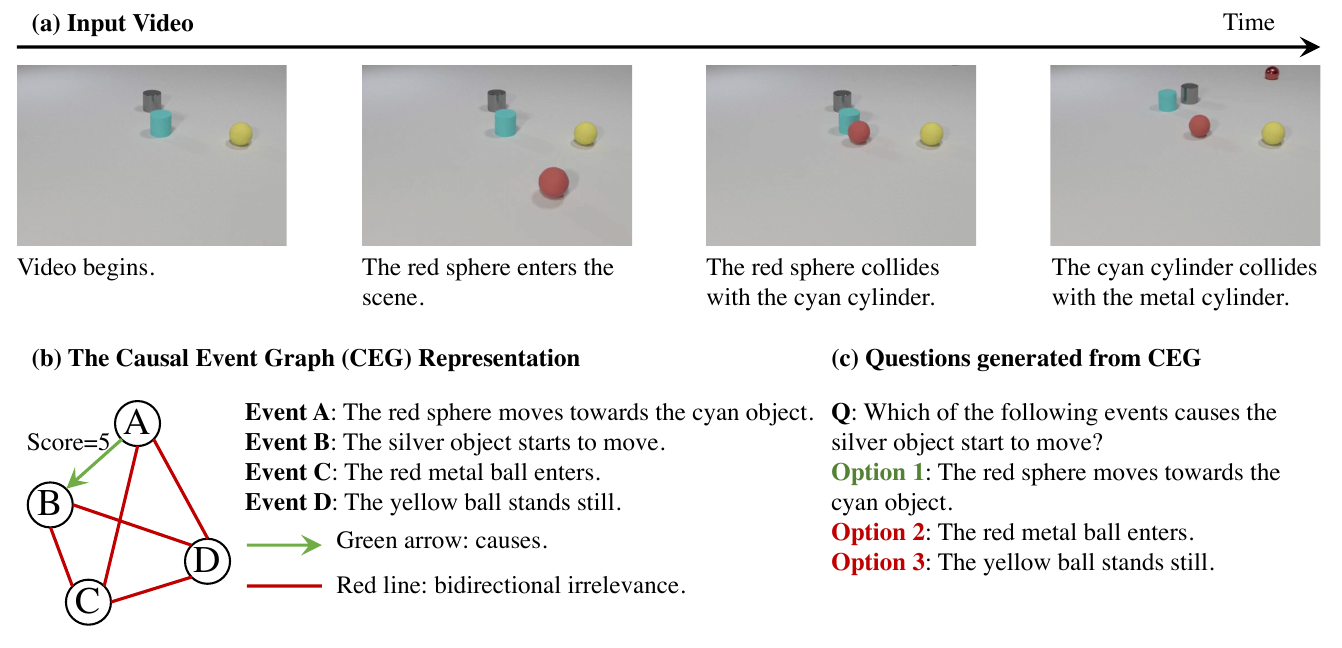}
    \caption{For (a) each video in the CLEVRER dataset, (b) \dataset annotates a human-labeled graphical representation of physical events and their causal relations, in the form of causal event graphs (CEGs). Each CEG composes of a collection of nodes associated with textual descriptions of events, and human-labeled directional edges indicating the causal relationship between objects. Each edge is also associated with a score indicating a human's graded attribute of causal relations. (c) The compact representation of CEGs can be easily translated into question-answer pairs to evaluate video reasoning models.}
    \label{fig:teaser}
\end{figure} 
We have seen significant progress toward machine reasoning about physical events and causal structures, driven by datasets such as CLEVRER~\citep{yi2019clevrer} and CATER~\citep{girdhar2019cater}. These datasets pair videos containing physical object interactions (\eg, collisions) with natural language question-answer pairs. However, there are two important design flaws in existing datasets. First, the categories of physical events are designed manually and identified through heuristic rules over object locations and velocities, significantly restricting the datasets' diversity. For example, the CLEVRER dataset only contains three types of events (object enter, object exit, and collision). Second, the causal relationships of events are also detected by heuristic rules. In contrast, human causal judgment resembles complex counterfactual reasoning processes~\citep{sloman2005causal,wolff2007representing,beller2020language}, and it is difficult to capture these subtleties through manually specified rules.

To address both shortcomings, we present CLEVRER-Humans, a human-annotated physical and causal reasoning dataset based on videos from CLEVRER~\citep{yi2019clevrer}, of which we show examples in \fig{fig:teaser}a. To ensure sufficient event diversity and annotation density (\ie, a dense causal relation between events), CLEVRER-Humans relies on a data representation termed Causal Event Graphs \citep[CEG; ][]{yi2019clevrer,ates2020craft}, whose nodes are natural language descriptions of the events in the video and edges are causal relationships between the events, as illustrated in \fig{fig:teaser}b. The CEG annotation procedure has two stages. In the first stage, we use an iterative event cloze task to collect event descriptions. Specifically, starting from a seeding set of events in the original CLEVRER dataset, we ask annotators to describe other events that are responsible for these seed events. The newly annotated events will be iteratively used as new seeds to progressively grow the annotated events. In the second stage, we condense the CEGs by asking human annotators to make binary classifications of causal influence for all pairs of physical events generated in the first stage. Based on both positive and negative labels, we can construct physical and causal reasoning questions in natural language, shown in \fig{fig:teaser}c.

One challenge of this pipeline is that human annotation for event cloze tasks is generally time- and budget-consuming. To alleviate this, we leverage the observation that neural generative models are reasonably good at generating event descriptions. Thus, we only collect a small number of physical event descriptions in the first stage. In the second stage, we augment these data by training neural event description models based on the ground truth physical trajectories of objects and ask human annotators to filter incorrect or hard-to-interpret descriptions.

Finally, we benchmark several machine learning models on CLEVRER-Humans. We demonstrate that our dataset is challenging, in particular, due to the diversity of event descriptions and the challenge of data-efficient learning. 
The development of \dataset can be beneficial to both machine learning and cognitive science communities. From the machine learning perspective, \dataset posits a combined challenge of natural language understanding, physical grounding of language, and causal reasoning in physical scenes. It is also a stimulus set for cognitive science studies of human physical event perception, causal judgment, and description~\citep{beller2020language,gerstenberg2022would}. %
\section{Related work}

\begin{table}[tp]
    \centering\small
    \setlength{\tabcolsep}{4.5pt}
    \begin{tabular}{lcccccc}
    \toprule
        Dataset & Video & \mycellc{Question\\Answering} & \mycellc{Diagnostic\\Annotation} & \mycellc{Natural Language\\Events} & \mycellc{Causal\\Reasoning} & \mycellc{Human Causal\\Judgements}\\ \midrule
        CLEVR~\citep{johnson2017clevr} & - & \checkmark & \checkmark & - & - & -\\
        MovieQA~\citep{tapaswi2016movieqa}  & \checkmark  & \checkmark & - & \checkmark & - & - \\
        TGIF-QA~\citep{jang2017tgif}        & \checkmark  & \checkmark & - & \checkmark & - & - \\
        TVQA+~\citep{lei2019tvqa}           & \checkmark  & \checkmark & - & \checkmark & - & - \\
        AGQA~\citep{grunde2021agqa}         & \checkmark  & \checkmark & - & \checkmark & - & - \\
        \midrule
        IntPhys~\citep{riochet2021intphys}& \checkmark  & - & \checkmark & - & \checkmark & -\\
        Galileo~\citep{wu2015galileo}     & \checkmark  & - & \checkmark & - & \checkmark & -\\
        PHYRE~\citep{bakhtin2019phyre}    & \checkmark  & - & \checkmark & - & \checkmark & -\\
        CATER~\citep{girdhar2019cater}    & \checkmark  & \checkmark & \checkmark & - & \checkmark & -\\
        CoPhy~\citep{baradel2019cophy}    & \checkmark  & - & \checkmark & - & \checkmark & -\\
        CRAFT~\citep{ates2020craft}       & \checkmark  & \checkmark & \checkmark & - & \checkmark & -\\
        CLEVRER~\citep{yi2019clevrer}     & \checkmark  & \checkmark & \checkmark & - & \checkmark & -\\
        ComPhy~\citep{chen2022comphy}     & \checkmark  & \checkmark & \checkmark & - & \checkmark & -\\ \midrule
        \dataset                & \checkmark  & \checkmark & \checkmark & \checkmark & \checkmark & \checkmark\\ \bottomrule
    \end{tabular}
    \vspace{4pt}
    \caption{Comparison between \dataset with other visual reasoning benchmarks. Our dataset is the only dataset of natural language descriptions of physical events and human judgments.}
    \label{tab:comparison}
\end{table}

\xhdr{Physical and causal reasoning.} Our dataset is closely related to benchmarks on physical reasoning tasks. In general, these benchmarks can be categorized into three groups based on their evaluation protocol. First, datasets such as IntPhys~\citep{riochet2021intphys}, Galileo~\citep{wu2015galileo}, and CoPhy~\citep{baradel2019cophy} focus on making counterfactual or hypothetical predictions of physical events. Second, benchmarks including PHYRE~\citep{bakhtin2019phyre} and ESPRIT~\citep{rajani2020esprit} focus on an ``inverse'' problem of generating initial conditions that leads to a specific goal state. The third group of benchmarks, where our dataset \dataset also falls into, focuses on assessing machine learning models through a natural language interface, such as 
CATER~\citep{girdhar2019cater}, CLEVRER~\citep{yi2019clevrer}, CRAFT~\citep{ates2020craft}, and ComPhy~\citep{chen2022comphy}. While these datasets focus on other aspects of physical and causal reasoning, such as physical property inference~\citep{wu2015galileo} and few-shot learning~\citep{chen2022comphy}, their causal relationships and event descriptions are all generated by synthetic rules. By contrast, our dataset contains human-annotated physical event descriptions and causal relationships.

\xhdr{Human causal cognition.} Understanding how humans perceive and reason about causal relationships has been a long-standing problem in cognitive science~\citep{sloman2005causal}. A number of theories have been proposed to model human causal cognition, such as Force Dynamics~\citep{talmy1988force,wolff2007representing}, Mental Models~\citep{khemlani2014causal}, Causal Models~\citep{goldvarg2001naive,sloman2009causal}, and Counterfactual Simulation~\citep{halpern2020causes,gerstenberg2021counterfactual}. These theories have a variety of aims, including causal induction, causal attribution, and grounding causal language such as "cause", "prevent", "help", "enable", and "allow". In this work, we study the relatively neutral causal connective "because", together with freely-generated descriptions of physical events.

Physical causation, in particular, has received much attention with rich experimental and theoretical literature. Among theories of physical causation judgements: Conserved Quantity Theory~\citep{dowe2000physical} predicts causal relations by inspecting how conserved quantities (\eg, momentum) transferred between events. Force Dynamics Theory~\citep[FDT;][]{talmy1988force,wolff2007representing} focuses on analyzing the mechanism of causation when bodies interact, using force vectors to represent the interaction between objects, and predicting judgments based on the direction and length of these vectors. Counterfactual Simulation Theory~\citep[CSM;][]{halpern2020causes,gerstenberg2021counterfactual} leverages an intuitive physics model to make counterfactual judgements (A causes B if A’s occurrence makes a difference to B’s occurrence). These theories often agree on predicted causal judgements, and are very often different from the simple heuristic used to annotate the "responsible for" relation of CLEVERER. There are also critical differences in predictions, particularly for complex events (where it is not always clear how to apply a given theory). Thus it is necessary to collect human judgements of the causal relation between events; in future work, it will be interesting to compare machine learning models derived from our data to psychological theories of physical causation.

\xhdr{Video question answering.}
The primary task we study is to answer questions about videos of physical scenes. Unlike benchmarks that focus on multi-modal learning in real-world videos about human-human and human-object interactions~\citep{tapaswi2016movieqa,jang2017tgif,lei2019tvqa,grunde2021agqa}, we focus on understanding physical events and their causal relationships. We summarize the key factors that differentiate our dataset from others in \tbl{tab:comparison}. Among them, \dataset is the only dataset that contains human-annotated physical events and causal judgements.

\xhdr{Video-conditioned text generation.}
Our dataset is built on recent advances in using deep neural networks to generate natural language descriptions of videos~\citep{krishna2017dense,zhou2018end,Luo2020UniVL,sun2019videobert}. A common approach to generating video event descriptions is to first apply per-frame neural network encoders to obtain per-frame features, and aggregate them over time. Popular aggregation algorithms include average pooling~\citep{venugopalan2014translating}, recurrent neural networks~\citep{donahue2015long}, and attention mechanisms~\citep{yao2015describing}. Our model for event description generation falls into the third group, and leverages temporal attention in generation.
\section{CLEVRER-Humans}
\vspace{-0.5em}

Our dataset is based on the videos from the CLEVRER dataset~\citep{yi2019clevrer}. Each video contains at most 5 colliding objects moving on a single plane in 3D. The objects have 3 different shapes (cylinder, cube, and sphere), 8 colors, and 2 materials (rubber and material). The questions in CLEVRER consists of four types: descriptive (``what color''), explanatory (``what’s responsible for''), predictive (``what will happen next''), and counterfactual (``what if''). In this paper, we particularly focus on explanatory questions, which evaluate machine reasoning about causal relationships. Explanatory questions query the causal structure of a video by asking for the cause of an event, providing multiple choices. There are 3 types of events defined in CLEVRER's explanatory questions: entering the scene, exiting the scene, and colliding with another object. In CLEVRER, the event descriptions are generated by pre-defined templates.

\vspace{-1em}
\subsection{Delving into Causal Relation Annotations}
\vspace{-0.5em}

\begin{figure}[!tp]
    \centering\small
    \includegraphics[width=1.0\textwidth]{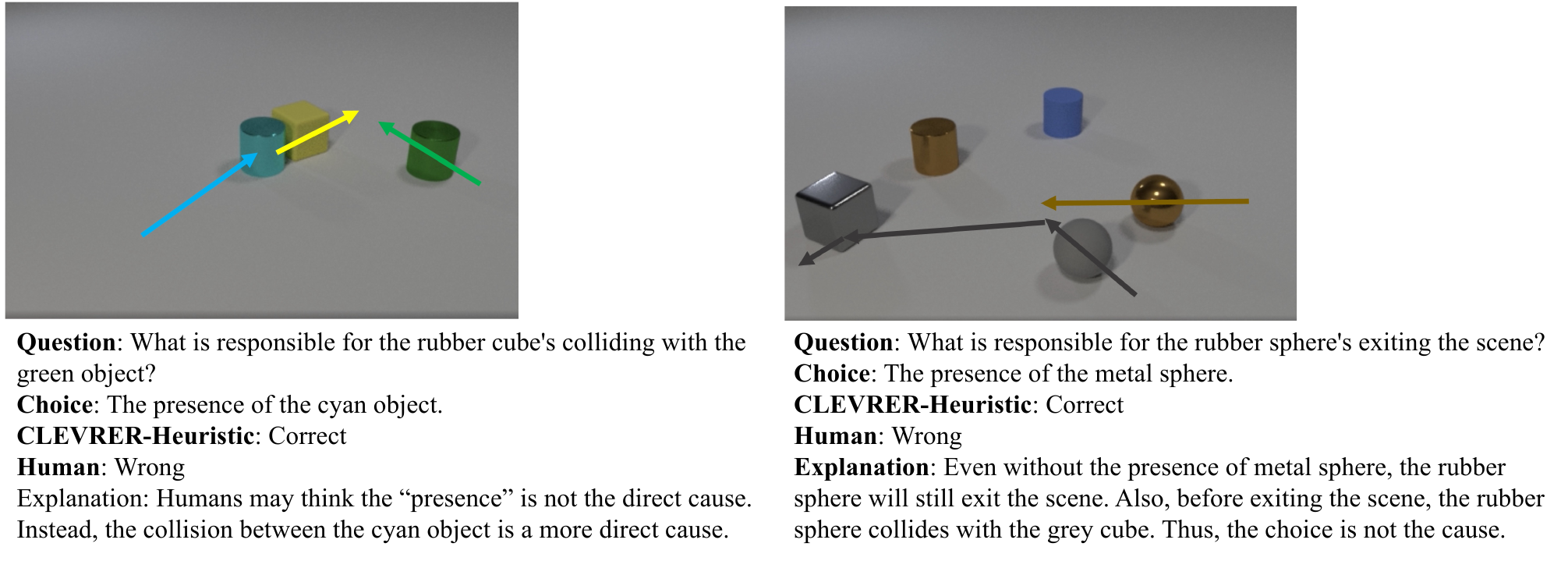}
    \caption{This are two examples showing difference of human causal judgment and CLEVRER's heuristic causal relation. The arrows in the image represent the moving direction of objects of interest.}
    \label{fig:causal_ablation}
\end{figure} 
\begin{table}[tp]
    \centering\small
    \setlength{\tabcolsep}{10pt}
    \vspace{4pt}
    \begin{tabular}{lccccc}
    \toprule
       P(X|Y) & Y = CLEVRER & Y = Counterfactual & Y = Human \\ \midrule
       X = CLEVRER & 1.00 & 0.74 & 0.96 \\
       X = Counterfactual &  0.89 & 1.00 & 0.54\\
       X = Human & 0.62  & 0.61 & 1.00 \\
        \bottomrule
    \end{tabular}
    \vspace{4pt}
    \caption{Comparison between different heuristics-generated causal labels and human labels, on a sampled subset of CLEVRER~\citep{yi2019clevrer}. The entry P(X|Y) denotes the fraction of event relations that are annotated as causal by protocol X given that the relations are annotated as causal by protocol Y.}
    \label{tab:ablation}
\end{table}

Before introducing our new dataset CLEVRER-Humans, we start from investigating the divergence between human labels and the labels generated by the original CLEVRER dataset using heuristics. In this section, we will be comparing three labelling protocols on a subset of 50 videos from the CLEVRER dataset. The first method (CLEVRER) uses the heuristic rules defined in the original CLEVRER dataset to predict causal relations between each pair of events. Specifically, if event A happens before event B, and they share at least one object, we say event A is a cause of event B\footnote{For simplicity, we use the word ``cause'' in this section. In the original CLEVRER dataset, such a relationship is formally defined as ``{\it A is responsible for B}.''}.  The second method (Counterfactual) uses counterfactual intervention to derive the causal relationship. Specifically, we say event A causes event B if event A happens before event B and event B happens even if we remove all relevant objects in event A from the scene (except for the objects in event B). We compare these two methods with human labeled causal relations.

We summarize the resulting statistics in Table \ref{tab:ablation}. There is a considerable difference between heuristic labels and human judgments. Compared to counterfactual intervention, the CLEVRER heuristic is a closer approximation to human judgment. We think this is partially because we have kept the term ``responsible for'' used by CLEVRER other than ``cause'' in the human labelling interface. \fig{fig:causal_ablation} provides two illustrative cases with explanations. We also provide detailed analysis on the effect of binarization thresholds and compositions of heuristics in the supplementary material.

\subsection{Augment CLEVRER with Human Annotations}
\begin{figure}[!tp]
    \centering\small
    \includegraphics[width=\textwidth]{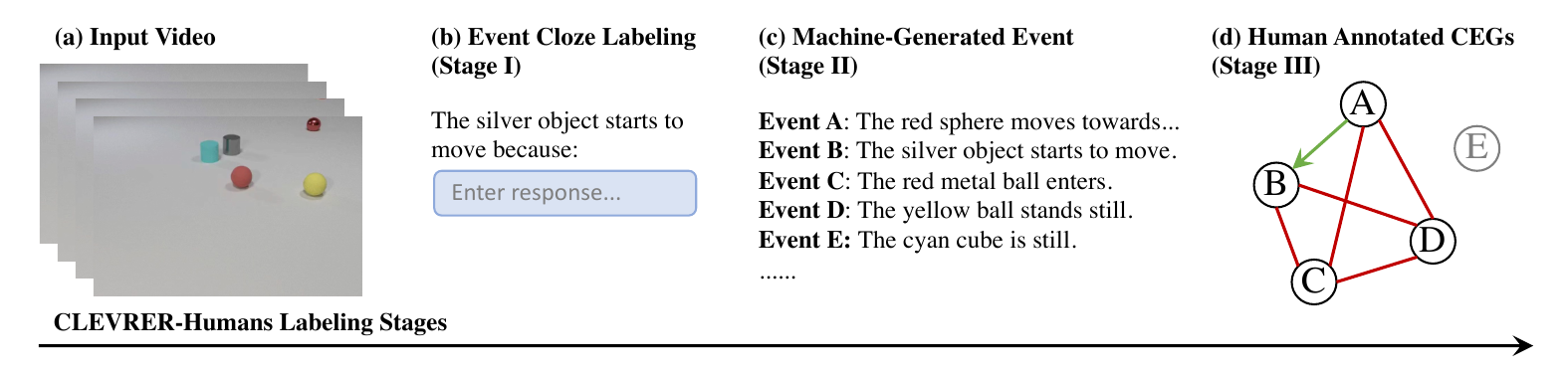}
    \caption{The overall labeling pipeline of \dataset. (a) Starting from input videos, (b) we use a event cloze task to collect a small number of human-written event descriptions about videos (Stage I). (c) Next, we train neural event description generators to augment all videos with a collection of events  (Stage II). (d) Finally, human annotators label the correctness of all generated events (in this case, event E is incorrect and thus the node is dropped) as well as their causal relations (Stage III).}
    \label{fig:pipeline}
\end{figure} 
The key representation we will be using to augment CLEVRER is the Causal Event Graph (CEG), which has been adopted in similar datasets on video causal reasoning such as CRAFT~\citep{ates2020craft} and video reasoning~\citep{irw_aaai21}. Each CEG $\gG = (\gV, \gE)$ is a graph structure of events and their causal relationships in a video. Each node $v \in \gV$ is a natural language description of a physical event in the video, and each directed edge $(v_1, v_2) \in \gE$ has a label in \{{\it positive, negative, unlabelled}\}\footnote{Note that CEGs are generally different from causal graphical models because edges in CEGs are manually labelled and they reflect human judgement. Thus, they may not comply inference rules in graphical models.}. A positive label for $(v_1, v_2)$ denotes that event $v_1$ is a cause of event $v_2$. CEGs directly enable downstream tasks such as physical and causal event generation, and causal relationship classification. Furthermore, based on the dense graph representation $\gG$, we can generate questions about the causal relations, \eg, ``Which event is responsible for the red cube moving?''

There are two desiderata for CEG annotations: diversity in event descriptions (nodes) and density in edge labels. Ideally, we want to ask human annotators to label all events in a video and relationships for all pairs of events. However, this is very time- and budget-consuming. Therefore, in practice, we use a three-stage annotation pipeline, as illustrated in \fig{fig:pipeline}. The first stage focuses on collecting human-written event descriptions using event cloze tasks, but only for a small number of videos. In the second stage, we augment the data for all videos using neural event description generators trained on the data collected from the first stage. In the third stage, we condense CEGs by collecting binary causal relation labels for all pairs of events from humans. Data are collected using the Mechanical Turk (MTurk) platform. 

\subsection{Stage I: Event Cloze}
\begin{wrapfigure}{r}{0.4\textwidth}
    \centering\small
    \vspace{-1.5em}
    \includegraphics[width=0.4\textwidth]{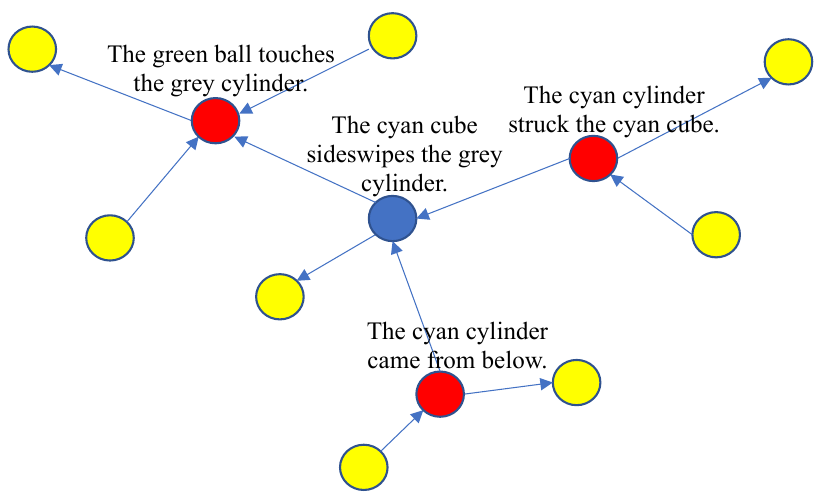}
    \caption{We use a novel iterative data collection procedure to collect CEGs on MTurk. Starting from a single node (iteration 0, blue), we iteratively sample nodes in the current CEG and collect either a cause or an effect event, and add this new node to the CEG. Red: nodes added in the first iteration; yellow: nodes added in the second iteration.}
    \label{fig:iterative_collection}
    \vspace{-4em}
\end{wrapfigure}

\begin{table}[tp] 
\centering\small
\begin{tabular}{L{4cm} L{4cm} L{4cm}}
\toprule
\textbf{Object State}                                                         & \textbf{Relative Position}           & \textbf{Collision}                                        \\ \bottomrule
Object A comes/moves/is hurled from some direction.                       & Object A is in the path of object B. & Object A is hit/struck by object B                        \\
Object A is thrown/pushed from some direction.                            & Object A is aimed at object B.       & Object A hits/collides with/bumps into/runs into object B \\
Object A changes direction (towards some direction)  &  Object A moves together with object B                                    & Object A is avoided hitting object B.                     \\
Object A stops moving  &                                      & Object A pushes object B into object C.                   \\ \bottomrule
\end{tabular}
\vspace{4pt}
\caption {Event examples collected in the first stage, categorized into single-object states, pairwise relative positions, and pairwise collision events.}
\label{table:event_type}
\end{table} 
In the first stage, we use an event cloze task to collect human-written event descriptions. Cloze tests have been employed in various natural language processing (NLP) domains, such as narrative cloze~\citep{chambers-jurafsky-2008-unsupervised}, reading comprehension~\citep{onishi2016large} and story-telling~\citep{mostafazadeh2016corpus}. In contrast to these existing work, we develop a novel {\it iterative} data collection procedure as shown in Figure~\ref{fig:iterative_collection}. For each video, we initialize its CEG with a single node using an event description from CLEVRER. Then we iteratively sample a node in the CEG, and ask humans to annotate a cause event or an effect event of the sampled event.

Specifically, the MTurk interface contains a single video and an incomplete sentence about two events in the video, connected by a ``because'' discourse marker: ``\_\_\_\_ because the yellow cylinder collides with the purple square.'' In this case, the human-written event is an effect of the event specified in the sentence. Similarly, we use the template ``Event A happens because \_\_\_\_.'' to collect cause events. Most notably, the newly collected events will be used as the seeding event in the next iteration. This progressive design thus improves the diversity and coverage of the collected event descriptions.

In this stage, we select 100 videos from the CLEVRER training set with the largest number of collisions since we want to work with videos with especially rich causal relationships. In total, we obtain 1,724 event descriptions from 100 CLEVRER videos through crowdsourcing. We recognize three major types of events in the stage I dataset as shown in Table~\ref{table:event_type}.
\subsection{Stage II: Trajectory-based Event Description Generation}

\begin{figure}[!tp]
    \centering\small
    \includegraphics[width=\textwidth]{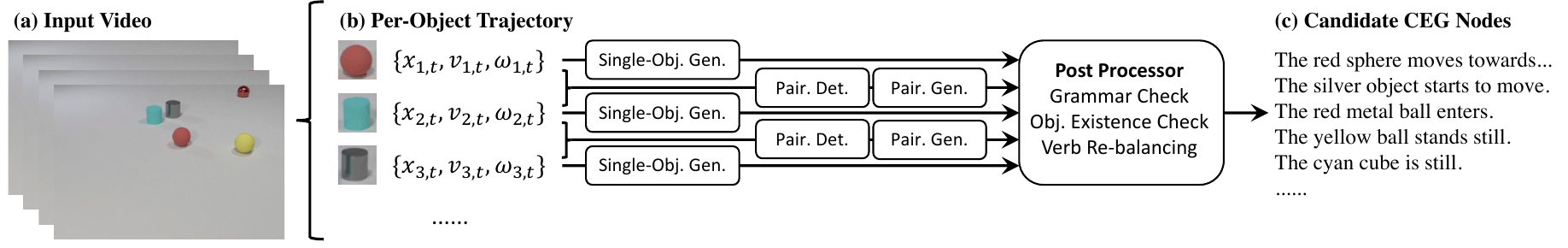}
    \caption{The neural event description generation pipeline (Stage II). (a) Given the input video, (b) we first extract per-object trajectories, composed of their attributes, positions, velocities, and angular velocities. For each object, we use a single-object event generator to sample event descriptions. For each pair of objects, we use a cascaded generator composed of a rule-based event detector and a neural pairwise generator. All generated events will pass a post-processing unit composed of three stages: grammar check, object existence check, and verb re-balancing. (c) The final product of the pipeline is a candidate node set of the CEG, which will be further annotated by humans in Stage III, the CEG condensation stage.}
    \vspace{-1em}
    \label{fig:generator}
\end{figure}

In the second stage, we leverage the manually written event descriptions to train neural description generators and augment the event description data for all videos. Our overall pipeline is shown in \fig{fig:generator}. Given the input video, it uses two branches to generate single-object events and pairwise events. The generated event descriptions will be sent to a post-processing module. It is important to note that the design choices in this stage focus on the {\it coverage} (\ie, we want to recover as many events from the input video as possible). All generated descriptions will be filtered again by human annotators.

\xhdr{Trajectory representation.} Instead of working with pixels, all event generation modules take symbolic representations of object trajectories as their inputs. Specifically, each video is represented by a set of trajectories $\{ s_{i,t} \}$, where $s_{i, t}$ is the state of object $i$ at timestep $t$. The state contains the color, shape, and material properties of object $i$, as well as its 3D position, velocity, and angular velocity at time step $t$. Since the CLEVRER videos are generated using simulated physical engines, this information can be directly obtained. Compared with images, trajectory-based inputs are lower-dimensional, and empirically yield significantly better results than pixel-based inputs when trained on a limited amount of data.

\xhdr{Data pre-processing.} To associate human-written event descriptions with a single object or a pair of objects, we first run noun-phrase detection methods\footnote{In particular, we used the tools from spaCy: \url{https://spacy.io/}.} on all descriptions. Next, we use all concept words in the detected noun phrases to filter the objects being referred to. The concept vocabulary contains colors, shapes, and materials, and is manually annotated. Based on the detected objects, we categorize each description into single-object (pairwise) events, and associate them with the corresponding object (pair of objects).

\xhdr{Single-Object event description.} The single-object event generator is a GRU-based sequence-to-sequence model with attention~\citep{cho2014gru}. It takes the trajectory of a single object throughout the video as input and generates natural language descriptions of events associated with this object. This module is trained on all of the 740 single-object descriptions from stage I.

\xhdr{Pairwise event description.} Unlike single-object events, an important feature of pairwise events is {\it sparsity}. Specifically, for most pairs of objects in the video, there is no event associated with them. However, because we don't have {\it negative} data in pairwise events (\ie, there are no descriptions such as ``{\it object A and object B do not have interactions.}''), direct training of pairwise event generators will yield a lot of false-positive detection during test time.

To address this issue, we employ a rule-based object pair filter before the event generation process. Concretely, the event detector takes in the trajectories of a pair of objects and outputs whether an event occurs for the input pair. It first detects the longest consecutive (increasing/decreasing) sequence of object positions to segment the input trajectories. Then it processes each segment based on the physical properties in the trajectory. Based on manual inspection of stage-I data, we use rules to detect three types of events: object approaching, collision, and moving together. This is a simple but effective pre-processing step. By choosing the threshold, our detector has a recall rate of 99.2\% on a small manually labeled dataset consisting of 100 videos. On the same split of 100 videos, the event detector improves the pairwise event description accuracy, labeled by human annotators in stage III, from 80.6\% to 87.2\%.
For all pairs selected by the pairwise event detector, we concatenate their trajectories and feed them into a separate GRU-based sequence-to-sequence model. Similar to the single-object generator, this model is trained on all of the 984 pairwise descriptions from stage I.

\xhdr{Post-processing.} We employ three post-processing steps to ensure the quality and diversity of generated event descriptions: grammar checking, object existence checking, and verb re-balancing. We first generate a large set of event descriptions (top-10 likely sequences per object and per pair). Next, we perform a grammar check using hand-crafted rules to filter common grammar errors such as missing verbs or missing verb arguments. Then, we detect noun phrases in the event descriptions and use concept words to select the object being referred to. We drop event descriptions that refer to objects that do not appear in the video. Finally, we notice that due to the sampling strategy, descriptions with the highest probabilities are not diverse: for example, for pairwise events, most highest-probability descriptions are about object collisions. Thus, we re-balance the data distribution based on the main verb of the descriptions. Specifically, we match this distribution with the human-written descriptions in stage I. For each video, we randomly sample at most 10 event descriptions.
\begin{figure}[tp]
    \centering\small
    \includegraphics[width=\textwidth]{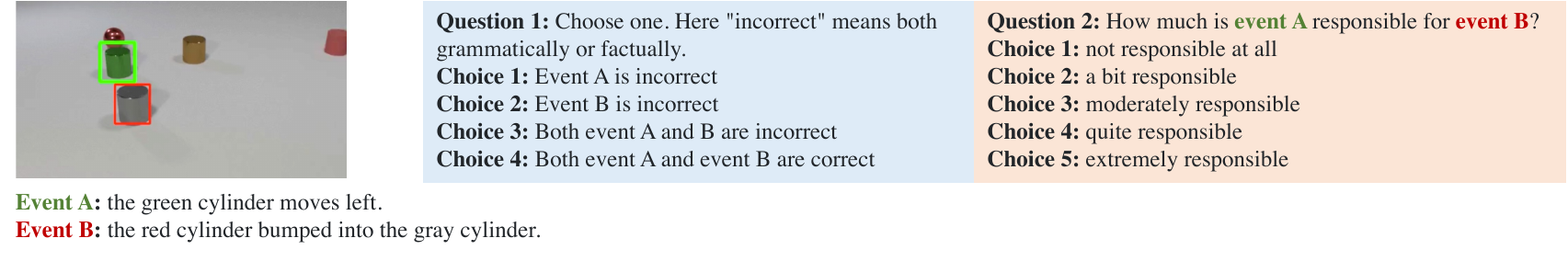}
    \vspace{-1em}
    \caption{The MTurk interface illustration for the second stage. The MTurker sees the input video with objects mentioned in the events highlighted with colored bounding boxes. The MTurker answers two questions about the video: 1) whether event descriptions are correct and 2) a graded judgement of whether event A is responsible for event B. We use the same wording choice ``be responsible for'' to be consistent with the original CLEVRER dataset~\citep{yi2019clevrer}.}
    \label{fig:stage2}
    \vspace{-1em}
\end{figure} 
\subsection{Stage III: CEG Condensation}
\vspace{-0.5em}
Finally, based on event descriptions generated in the second stage, we build dense CEGs using MTurk. In contrast to the first stage, where we use an event cloze task to collect natural language descriptions, in this stage, we focus solely on edge labeling: classifying whether two events have a causal relation.

Concretely, we use an interface depicted in \fig{fig:stage2}. The annotator is presented with a video, with objects being referred to in event descriptions highlighted in different colors. The annotator needs to answer two questions. The first question asks whether two event descriptions are interpretable (\eg, grammatically correct) and correct (\ie, they happen in the video). If the annotator labels both descriptions correct, they need to answer a second question, asking for a graded judgment (score 1-5) about whether the first event is a cause of the second event.

Based on the responses, we drop CEG nodes where a majority of the annotators label incorrect. The labeled edges have a score ranging from 1 to 5. We binarize the labels into positive and negative edges with a threshold of 4. 
We also manually filtered out ambiguous event descriptions and enriched the negative edges (score $\le$ 3) in the CEG data files. 

\subsection{QA conversion}
\vspace{-0.5em}

To be compatible with the existing dataset CLEVRER, we further convert the CEGs to multiple-choice question-answer pairs. For every node $v$ in the CEG $\gG$ with an adequate amount of positive parents (\ie. ``causes''), we uniformly sample the number of correct choices in the question. We convert it to a question ``Which of the following is responsible for $A$?'' and sample the set of positive and negative answers following the practice of CLEVRER. This ensures that the distribution of positive and negative candidates is balanced, as in the original CLEVRER dataset. We use the same train, validation, and test splits as CLEVRER.

\subsection{Dataset Statistics} 
\vspace{-0.5em}

Overall, in \dataset, we retrieved ${1099}$ videos with ${5538}$ descriptions and ${23032}$ event relationship annotations, after dropping empty CEGs. Based on the CEGs, we generate ${1413}$ question-answer pairs. We will discuss the statistics of \dataset in detail in the supplementary material. In this section, we summarize four important features of \dataset.

First, \dataset contains dense annotations of causal relations between physical events. The average number of CEG nodes is ${4.88}$, and the average number of labeled edges is ${20.68}$. These dense annotations of CEGs form the rich and complicated causal structures in our dataset. Second, CLEVRER-Humans offers diverse free-form language descriptions while retaining balances in object properties.  CLEVRER-Humans has a vocabulary length of size ${157}$, which is much greater than CLEVRER (82). Next, most importantly, CLEVRER-Humans engage in a variety of physical events for causal reasoning tasks. In particular, \dataset contains ${30}$ distinct verbs, and verbs are used in different tenses. In comparison, the original CLEVRER dataset contains only three event types (and verbs): enter, exit, and collide. Therefore, CLEVRER-Humans significantly improves diversity and brings in a challenge for machines to recognize and ground these events in practice. Finally, CLEVRER-Humans' annotation reflects the subjective judgment of causality in physical events. CLEVRER-Humans offers 5 choices when asking MTurkers to label the causality level. The average score is ${1.70}$. Note that this distribution is skewed towards lower scores. This reflects the fact that most event pairs do not have causal relationships. Finally, although we have binarized the edge labels for the sake of consistency with CLEVRER, the raw score-based judgment can be potentially helpful in other tasks, such as cognitive science studies.

Therefore, we can conclude that CLEVRER-Humans is a high-quality causal relation dataset with significantly more diverse event types and language descriptions than CLEVRER.
\section{Experiments}
\vspace{-0.5em}

In this section, we present the question-answering results of a collection of baseline methods on our new dataset \dataset. Since we have adapted the same input-output interface as the original CLEVRER dataset~\citep{yi2019clevrer}, most methods that are applicable to CLEVRER can also be applied to our dataset. In particular, we compare the following representative ones and highlight the challenge of our diverse and human-annotated physical and causal reasoning task.

\subsection{Methods}
\xhdr{Per-option best guess.} The most basic baseline, per-option random guess (Guess) classifies each option with the most-frequent answer (``No'' in our dataset).

\xhdr{Language-only models.} We use language-only models to test potential language biases in the dataset. Specifically, we use an LSTM encoder~\citep{hochreiter1997long} to encode the natural language question and the option, concatenate the last hidden state of both sequences, and apply a linear layer for binary classification.

\xhdr{Program-based video QA models.} We also compare our model with NS-DR~\citep{yi2019clevrer} and VRDP~\citep{zfchen2021iclr}, two state-of-the-art video reasoning models based on program representations of questions. In particular, both methods leverage a semantic parser to parse questions and options into symbolic programs with hierarchical structures and execute the program based on the abstract video representation extracted by an object property network (NS-DR), or neuro-symbolic concept learning modules (VRDP). We train the semantic parser on the original CLEVRER dataset using the ground truth program annotations and use it to parse newly-annotated \dataset questions.

\xhdr{End-to-end video QA models.} We also evaluate three additional end-to-end video QA models: CNN+LSTM, CNN+BERT, and ALOE~\citep{ding2020object}. In CNN+LSTM, we use CNNs to encode each frame into latent vectors and use two separate LSTMs to encode the video sequence and questions, respectively. The output of both video LSTM and the question LSTM are concatenated and fed into another linear layer to make binary classifications. The CNN+BERT model uses a similar architecture except that we replace the LSTM encoder with a pretrained BERT encoder \citep{devlin2018bert}. ALOE~\citep{ding2020object} is the state-of-the-art video reasoning model on CLEVRER. It is based on the Transformer architecture~\citep{vaswani2017attention}. It uses MoNET~\citep{burgess2019monet} to extract an object-centric representation of videos, uses a multi-modal transformer to encode both object features and questions, and predicts the binary label.

\subsection{Results}
\begin{table}[tp]
    \centering\small
    \setlength{\tabcolsep}{9pt}
    \begin{tabular}{llcccc}
    \toprule
        & & \multicolumn{2}{c}{CLEVRER} & \multicolumn{2}{c}{\dataset} \\
        Model & Training & Per-Option & Per-Ques. & Per-Option & Per-Ques. \\ \midrule
        Best Guess     & N/A     & 50.2 & 16.5 & 50.7 & 30.1 \\
        Lang-Only & Scratch & 59.7 & 13.6 & 51.9 ($\pm$ 1.09) & 30.4 ($\pm$ 1.90) \\ \midrule
        NS-DR~\citep{yi2019clevrer} & Pretrain & 87.6 & 79.6 & 51.0 & 32.0 \\
        VRDP~\citep{ding2021dynamic}& Pretrain & 96.3 & 91.9 & 50.9 & 31.6 \\ \midrule
        CNN+LSTM                    & Pretrain & 62.0 & 17.5 & 50.3 & 30.0 \\
        CNN+LSTM                    & Scratch  & N/A$^\dagger$& N/A$^\dagger$ & 51.7 ($\pm$ 0.64) & {\bf 34.2} ($\pm$ 1.69) \\
        CNN+LSTM                    & Pretrain+Finetune & 62.0 & 17.5 & 51.5 ($\pm$ 2.35) & 30.8 ($\pm$ 0.69)\\
        CNN+BERT  & Pretrain & 55.1 & 11.5 & 52.9 & 32.0 \\
        CNN+BERT  & Scratch  & N/A$^\dagger$& N/A$^\dagger$ & 52.0 ($\pm$ 2.34)  &  30.2 ($\pm$ 2.41)\\
        CNN+BERT  & Pretrain+Finetune  & 55.1 & 11.5 & 50.1 ($\pm$ 0.68)  &  30.4 ($\pm$ 3.09)\\
        ALOE~\citep{ding2020object} & Pretrain & {\bf 98.5} & {\bf 96.0} & {\bf 54.0} & 26.9 \\
        ALOE~\citep{ding2020object} & Scratch  & N/A$^\dagger$& N/A$^\dagger$ & 51.8 ($\pm$ 1.00) & 31.7 ($\pm$ 0.79) \\
        ALOE~\citep{ding2020object} & Pretrain+Finetune  & {\bf 98.5} & {\bf 96.0} & 52.7 ($\pm$ 1.36) & 32.1 ($\pm$ 1.36) \\ \midrule
        Human                       & N/A                & N/A        & N/A        & 84.5 & 71.4 \\

        \bottomrule
    \end{tabular}
    \vspace{10pt}
    \caption{Test performance of different models on both the original CLEVRER dataset and our new \dataset dataset. The training column denotes the training schema for different models (in percentage, and the $\pm$ sign shows the 95\% confidence interval computed across 5 different runs). We compare both per-option accuracy and per-question accuracy, following the original paper~\citep{yi2019clevrer}. Note that the number of options per question is $\sim$4 for CLEVRER and 2 for CLEVRER-Humans. N/A marker$^\dagger$: models trained only on CLEVRER-Humans.}
    \label{tab:results}
    \vspace{-1em}
\end{table}

Our results are summarized in \tbl{tab:results}. For both program-based models, since we do not have program annotations for \dataset questions and options, they are trained only on the original CLEVRER dataset and tested on \dataset. For both end-to-end models, we compare three alternatives: trained on CLEVRER, trained on \dataset, and pretrained on CLEVRER and then finetuned on \dataset. For most models that are pretrained, we used the checkpoints released by the authors. Thus, we are unable to compute confidence intervals.

Overall, all methods perform poorly on our dataset \dataset, especially when compared with the best guess model. In general, these results highlight three distinctive challenges of \dataset. First, the diversity of events: the vocabulary of \dataset questions and options are significantly richer than the original CLEVRER dataset. As a result, the performance of models pretrained on CLEVRER significantly drops (compared to their CLEVRER performances). In particular, the original CLEVRER dataset only has a vocabulary size of 82, while \dataset has a vocabulary size of 219. To directly apply pretrained models, we have to encode a large portion of textual inputs as ``out-of-vocabulary.'' Furthermore, by comparing CNN+LSTM and CNN+BERT, we see that using pretrained language models is not necessarily helpful for generalization to unseen events. The second challenge that accounts for the performance gap between the original CLEVRER and our dataset is human-annotated causal judgments: we do not see a significant difference between models trained from scratch on \dataset and the ones pretrained on CLEVRER in terms of the final performance, which empirically suggests the gap between causal relation labels between two datasets. A third challenge that arises from training-from-scratch or finetuning is the limited size of the \dataset training set. Recall that our dataset is directly comparable with the ``explanatory questions'' category of CLEVRER. CLEVRER contains 122,461 explanatory question pairs, while \dataset contains 1076 pairs. As a result, we observe extreme training-set overfitting for large models such as ALOE. While as training goes on, the training accuracy keeps increasing to 70\%, the per-option test accuracy plateaus at 53.7\% at a very early stage. The combined challenge of diversity and data-efficient learning can be potentially addressed by enabling better transfer learning from large pre-trained multi-modal models such as CLIP~\citep{radford2021learning}, and better physics-informed models~\citep{wu2017learning}. We also include human performance on the test set. Participants with full professional or higher proficiency in English are asked to evaluate the results on 50 videos from the test set. The participants also provide the percentage of descriptions that appear natural to them, which is 90.0\% on average.

\vspace{-0.5em}
\section{Conclusion}
\vspace{-0.5em}

We have presented \dataset, the first video reasoning dataset of human-annotated physical event descriptions and their causal relations. \dataset introduces a unique and important challenge of combined physical scene understanding, natural language understanding, and causal reasoning. Due to its limited size, CLEVRER-Humans should be primarily used for zero-shot evaluation or few-shot training. Our preliminary results on question-answering illustrate the challenge of interpreting diverse human-written event descriptions, making human-like causal judgments, and data-efficient learning. To collect the dataset within a reasonable budget, we introduce two important techniques for dataset collection: iterative cloze-based annotation of event descriptions, and hybrid event description generation using neural sequence-to-sequence models. Both techniques can be applied to elicit data for any similar relation (such as ``before'' and ``supports''). 
An exciting challenge will be extending these methods to naturalistic videos, complex events, and ambiguous relations.

Throughout the evaluations in this paper, we have been focusing on a binarized version of the causal reasoning task. However, it is important that computational models can make graded judgements as humans do. While existing literature has studied graded causal reasoning in text domains \citep{roemmele2011choice,zhang2017ordinal} and static images~\citep{yeo2018visual}, we believe an important direction is to build better models that perform graded causal reasoning in dynamic videos.

\xhdr{Acknowledgements.} This work is partly supported by the Stanford Institute for Human-Centered AI (HAI), the Samsung Global Research Outreach (GRO) Program, ONR MURI N00014-22-1-2740, and Adobe, Amazon, Analog, Bosch, IBM, JPMC, Meta, and Salesforce. We also thank Keyi Hu for her contribution.

{\small
\bibliographystyle{unsrt}
\bibliography{reference}
}

\clearpage
\appendix

\appendix
We bear all responsibility in case of violation of rights. The data created or used during this study are openly available on the project's website (\url{https://sites.google.com/stanford.edu/clevrer-humans}). We confirm that the data is under a CC0 license. We will provide maintenance to the website and dataset regularly and upon request.

The rest of this supplementary document is organized as the following. First, in \sect{app:visualizations} we provide visualizations of data collected in \dataset, as well as more analysis on the comparison between human causal judgments and heuristics-based labels. In \sect{app:implementation}, we describe the implementation details of models studied in the main paper and add additional failure case analysis of models. Next, in \sect{app:interface}, we describe the user interface for dataset collection. Finally, in \sect{app:sheet}, we supplement dataset sheets for \dataset.

\section{Dataset Visualization and Analysis}
\label{app:visualizations}
\begin{figure}[!tp]
    \centering\small
    \includegraphics[width=\textwidth]{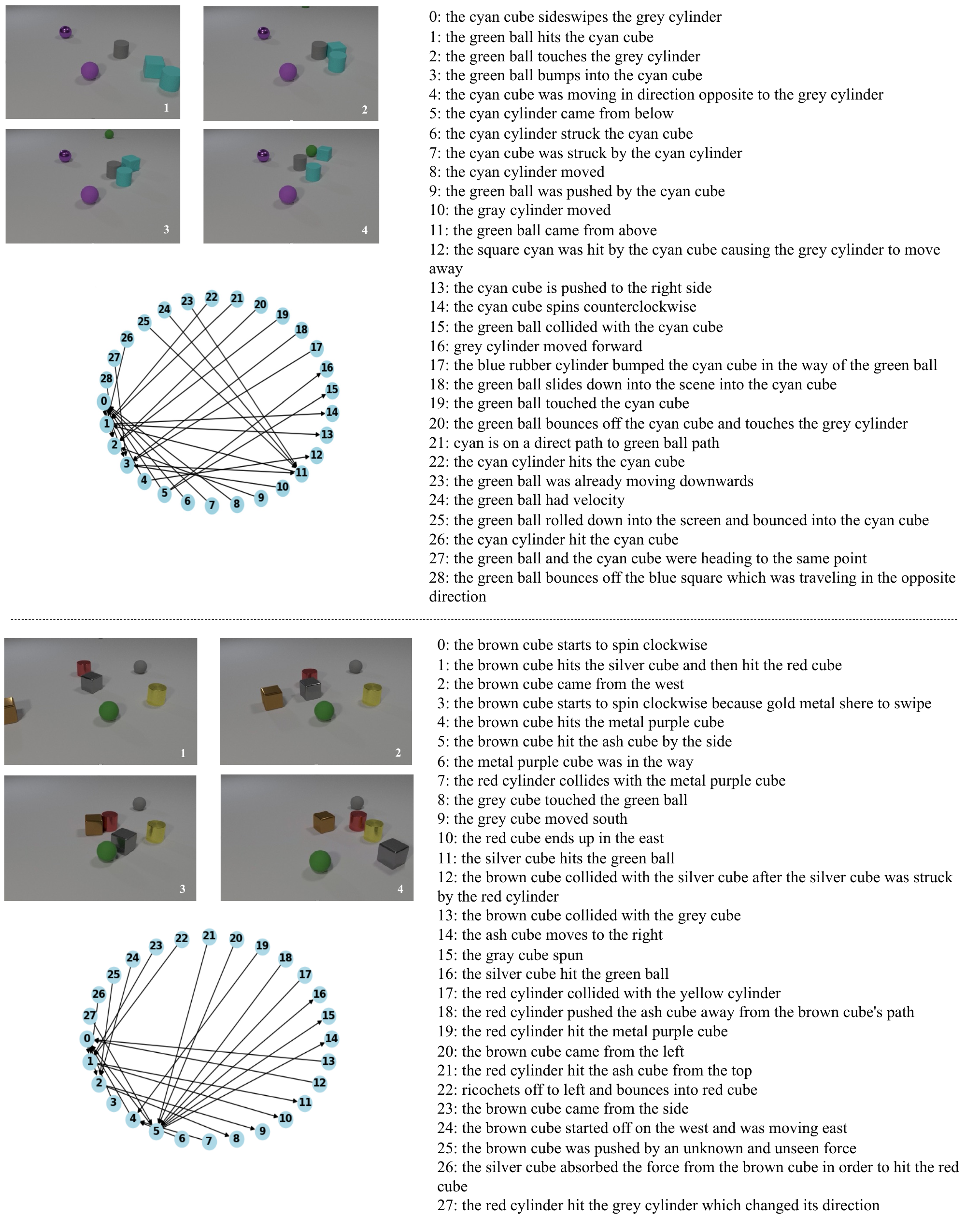}
    \vspace{-1em}
    \caption{Visualization of two samples of annotations collected in Stage I causal cloze tasks. They are collected progressively by feeding the response of a user as the input of another one. The black arrows indicate the annotation orders. }
    \label{fig:graph_example_cloze}
    \vspace{1.5em}
\end{figure}

\begin{figure}[!tp]
    \centering\small
    \includegraphics[width=0.7\textwidth]{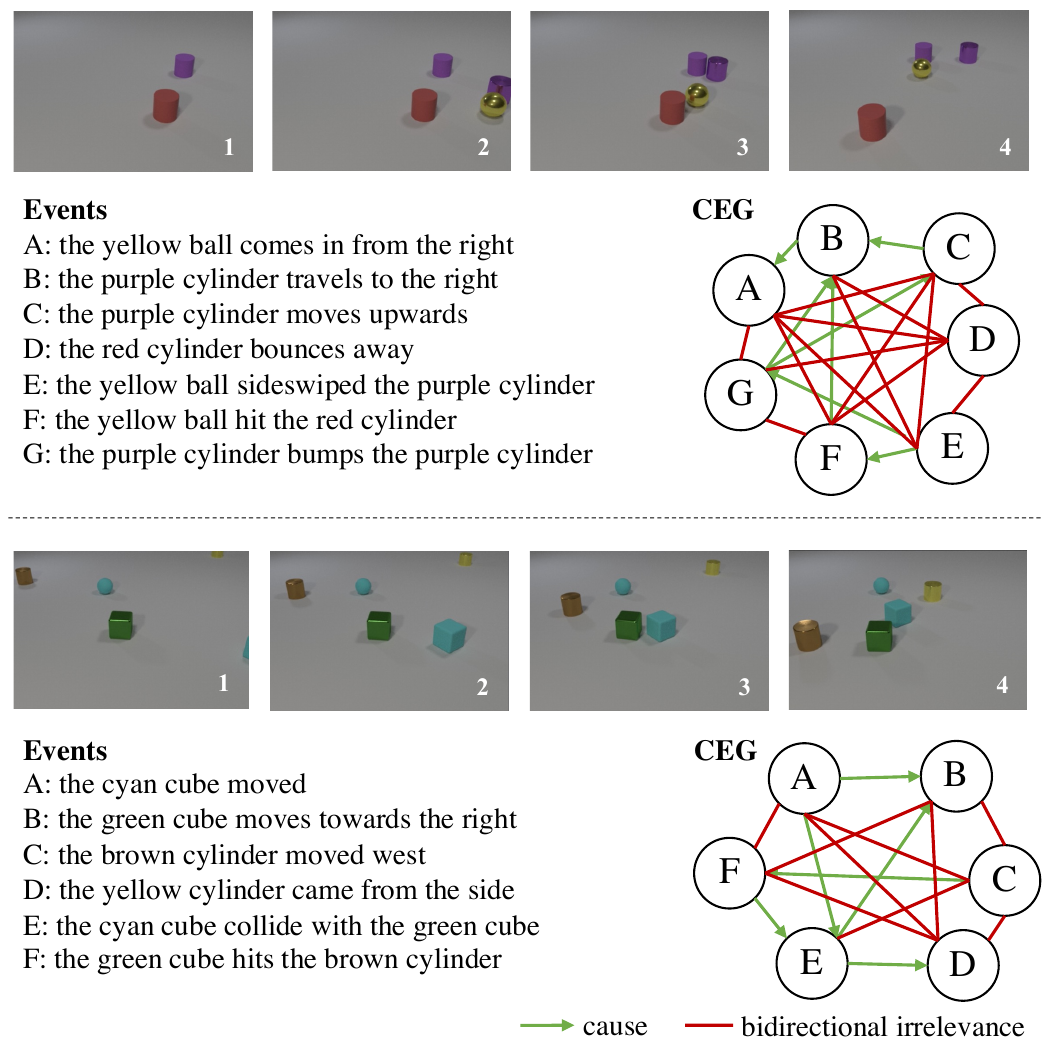}
    \vspace{0em}
    \caption{Visualization of two samples of CEGs in \dataset. The green arrows represent causal relations and the red edges represent bidirectional irrelevance. We can see the rich causal relations among physical events presented in the CEGs.}
    \label{fig:graph_example}
    \vspace{1.5em}
\end{figure}

\fig{fig:graph_example_cloze} and \fig{fig:graph_example} show the example graph collected in the stage I (causal event cloze) and stage II (binary CEG labelling), respectively. First, \fig{fig:graph_example_cloze} shows that the causal cloze tasks can progressively collect a large number of human-written event descriptions by re-using the response of previous annotators. On average, we can obtain 29.4 descriptions per video, highlighting the advantage of our design. Second, the condensed CEGs contain high-quality causal relations of physical events, as shown in \fig{fig:graph_example}. It demonstrates both the language diversity and the richness of causal relations in the CEGs of \dataset. These figures provide a straightforward illustration of our data collection pipeline and the quality of our data.

\vspace{-0.5em}
\subsection{Dataset Statistics} 
\vspace{-0.5em}

\begin{figure}[!tp]
    \centering\small
    \includegraphics[width=\textwidth]{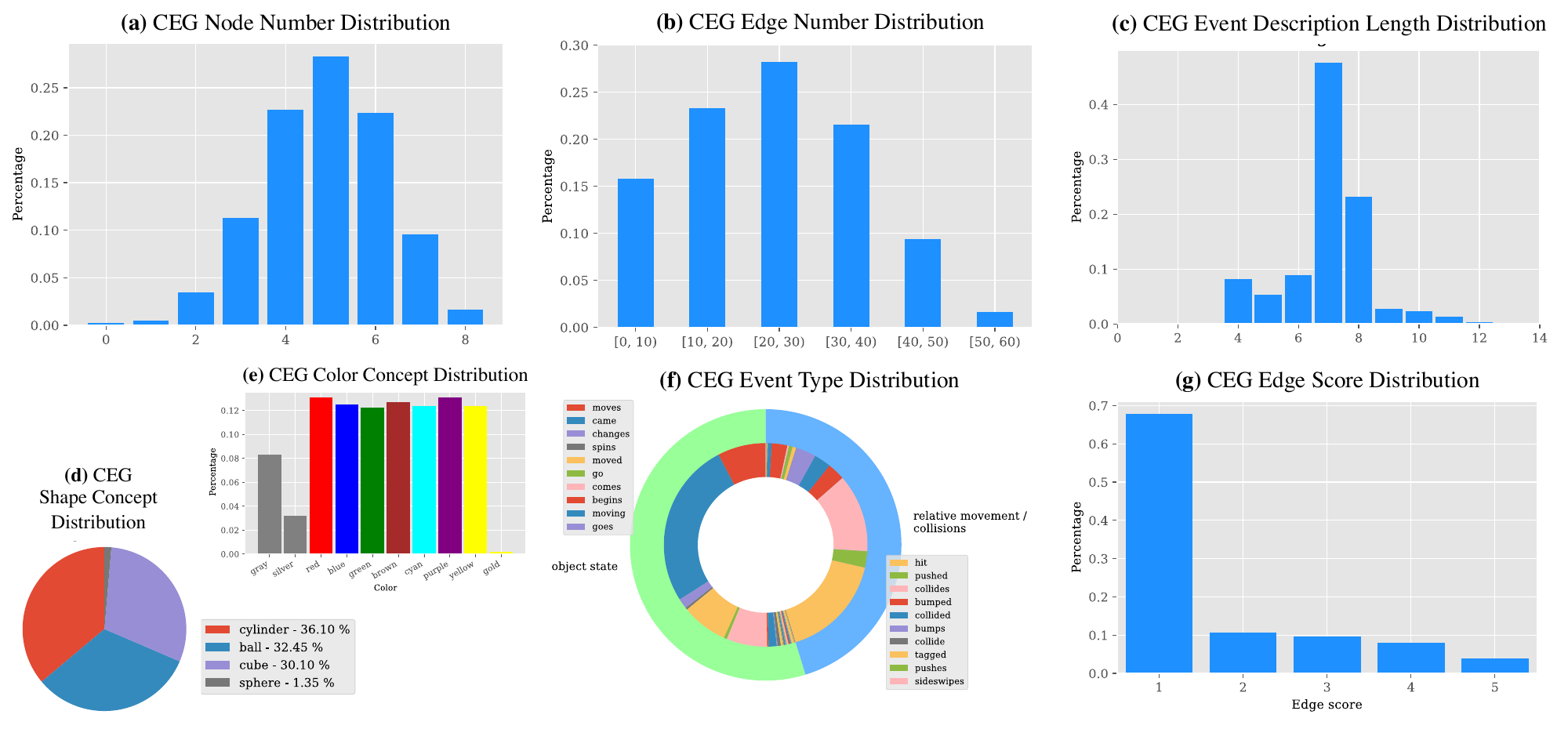}
    \vspace{-1em}
    \caption{Statistics on the CLEVRER-Humans dataset. From left to right, the first row figures are distributions of (a) the number of nodes per CEG, (b) the number of edges per CEG, and (c) sentence lengths excluding the "which of the following is responsible for" prefix. The second row figures are distributions of (d) object shapes, (e) colors, (f) event type attributions based on verbs, and (g) CEG edge scores labelled by MTurkers, respectively.
    }
    \label{fig:dataset_stats}
    \vspace{-1.5em}
\end{figure}
 
First, \dataset contains dense annotations of causal relations between physical events. \fig{fig:dataset_stats}a and \fig{fig:dataset_stats}b show the distributions of the number of nodes and edges in each CEG. The average number of CEG nodes is ${4.88}$, and the average number of labeled edges is ${20.68}$. These dense annotations of CEGs form the rich and complicated causal structures in our dataset. 

Second, CLEVRER-Humans offers diverse free-form language descriptions while retaining balances in object properties. \fig{fig:dataset_stats}c shows the length distribution of event descriptions: the average length is ${6.99}$ (as a reference, the average event description length of CLEVRER is $8.93$). CLEVRER-Humans has a vocabulary length of size ${157}$, which is much greater than CLEVRER (82). \fig{fig:dataset_stats}d and \fig{fig:dataset_stats}e show the distribution of object property concepts: colors and shapes. They remain unbiased when considering the synonyms such as ``ball'' and ``sphere'' and ``gray'' and ``silver.''

Next, most importantly, CLEVRER-Humans engage in a variety of physical events for causal reasoning tasks. In particular, \fig{fig:dataset_stats}f shows the distribution of event types computed based on the main verb of the event description. The outer circle represents the general event families. The corresponding inner breakdowns display more than 10 variations of the expression based on verbs for each event type. In comparison, the original CLEVRER dataset contains only three event types (and verbs): enter, exit, and collide. Therefore, CLEVRER-Humans significantly improves diversity and brings in a challenge for machines to recognize and ground these events in practice.

In the following box, we list all verbs that have been annotated by human annotators and generated by our machine generative model. We have lemmatized all verbs to remove the tense.

\begin{shaded}
\begin{adjustwidth}{2em}{2em}
move, come, change, spin, move, %
go, begin, slow, travel, bounce, %
slide, stop, get, dash, run, roll, %
stand, leave, halt, lose, hurl, %
hit, push, collide, bump, tag, %
sideswipe, propell, strike, strike %
\end{adjustwidth}
\end{shaded}

We also would like to point out that for some verbs, if they seem to be synonyms (\eg, bump and sideswipe), they can have subtle differences in physical grounding. For example, A bumping into B usually implies that A is moving faster than B and its collision changed the state of B. Furthermore, the different tenses of the same verb have different meanings in sentences: "the event that ball A moved is responsible for the collision" is different from "the event that ball A is moving is responsible for the collision." In the former case, ball A does not have to be moving while the collision happens. 

It is possible to hand-craft a lot of rules to handle each individual case (\eg, bump, sideswipe, roll), but that will require additional hyperparameters for thresholds, and may be hard to align with human perception.

Finally, CLEVRER-Humans' annotation reflects the subjective judgment of causality in physical events. CLEVRER-Humans offers 5 choices when asking MTurkers to label the causality level. \fig{fig:dataset_stats}g shows the distribution of edge scores with an average of ${1.70}$. Note that this distribution is skewed towards lower scores. This reflects the fact that most event pairs do not have causal relationships. Finally, although we have binarized the edge labels for the sake of consistency with CLEVRER, the raw score-based judgment can be potentially helpful in other tasks, such as cognitive science studies.

Therefore, we can conclude that CLEVRER-Humans is a high-quality causal relation dataset with significantly more diverse event types and language descriptions than CLEVRER.

\begin{figure}[!tp]
    \centering\small
    \vspace{-0.5em}
    \includegraphics[width=0.5\textwidth]{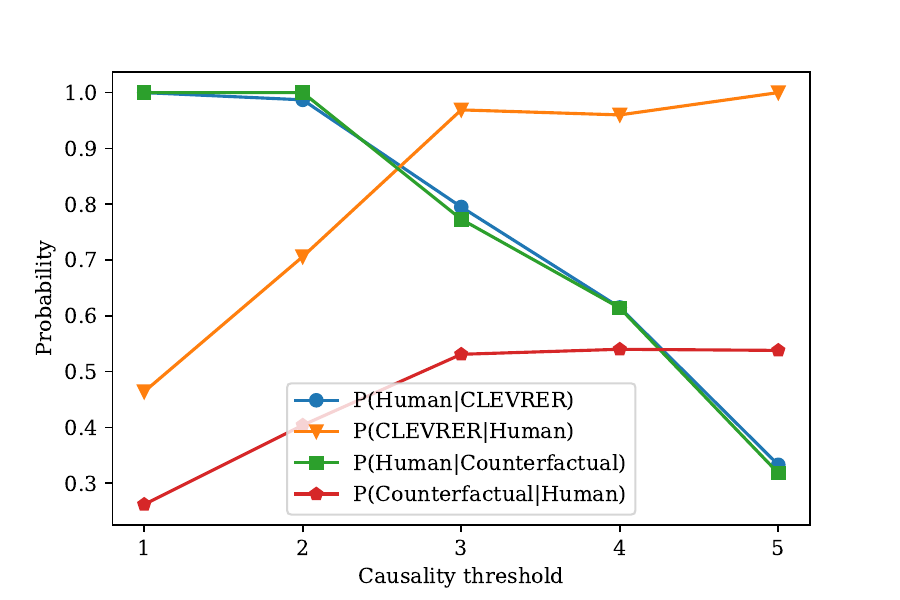}
    \caption{Effect of different causality thresholds on the binarized human causal relation. The x-axis is the ablation threshold (i.e., 4 means a score $\geq$ 4 represents a causal relation). The y-axis is the conditional probability.}
    \vspace{-1.5em}
    \label{fig:ablation_threshold}
\end{figure} \begin{table}[tp]
    \centering\small
    \setlength{\tabcolsep}{10pt}
    \vspace{4pt}
    \begin{tabular}{lccccccc}
            \toprule
        & Y = $y_1$ & Y = $y_2$ & Y = $y_1 \wedge y_2 $  & Y = $y_1 \vee y_2 $ & Y = $y_1 \oplus y_2 $\\ \midrule
    
       P(X = Human | Y) & 0.62 & 0.61 & 0.23  & 0.34 & 0.34\\
        P(Y | X = Human) & 0.96 & 0.54 & 0.29 & 0.62 & 0.33\\
        \bottomrule
    \end{tabular}
    \vspace{4pt}
    \caption{Comparison between different combinations of heuristics-generated causal labels and human labels, on a sampled subset of CLEVRER~\citep{yi2019clevrer}. The entry P(X|Y) denotes the fraction of event relations that are annotated as causal by protocol X given that the relations are annotated as causal by protocol Y. $y_1, y_2$ denote the existence of causal relations defined CLEVRER's heuristic and Counterfactual causal relation, respectively.}
    \label{tab:ablation_combination}
    \vspace{-2em}
\end{table}

\subsection{Comparison between Heuristic and Human Causal Judgments}
We supplement the effect of different thresholds on the graded causal relation in \fig{fig:ablation_threshold}. In the human performance study, we asked the participants to choose a threshold from 1-5 if they had to binarize their judgment. The average threshold suggested by the participants is 3.6. In practice, we choose a threshold of 4 to obtain the causal relation that humans are more certain about.

Having shown the two common heuristics-generated causal labels (CLEVRER's and counterfactual intervention) diverge from human judgment, we also provide the results on comparisons between different combinations of heuristics-generated causal labels and human judgments. We use the logic operators and ($\wedge$), or ($\vee$), xor ($\oplus$). As shown in Table \ref{tab:ablation_combination}, none of these combinations can give a close enough approximation to human judgment, which further justifies our motivation to use human-labeled causal data for \dataset.

\clearpage

\section{Implementation Details}
\label{app:implementation}

In this section, we present the implementation details of our neural network-based event description generator, the baseline models studied in the main paper, and the error bars for models across different random seeds.

\subsection{Stage II Implementation}
We first describe the input pre-processing for neural event generators. For each object, we concatenate the one-hot encoding of physical properties (including shapes, colors, and materials) and the motion information (including location, orientation, velocity, angular velocity, and whether the object is inside the camera view) in each of the 128 frames in a video. For each object, at each time step, the input dimension is 24. 

Our rule-based event detector for object pairs works as the following.
For object pairs, we first extract all segments that are composed of consecutive frames when two objects are close two each other. Specifically, we say two objects are close if the $L_\infty$ norm of the displacement vector between two objects is smaller than 0.5 meters (\ie, their x, y, z displacements are all smaller than 0.5 meters). Within each segment, the event detector predicts event types, including moving together, object approaching, and collision, based on changes in the motion information. For example, if two objects are physically close for more than 20 frames without rapid changes in velocity, we consider them relatively static, thus ``moving together.'' If both objects change directions within their close period, we consider a collision happened. We can further distinguish the changes in relative positions (either ``bouncing back'' or ``one approaching another'') by the sign of the dot product of velocity vectors. For any object pair, if no events are detected in the course of the entire video, we do not include this pair for future captioning. 

After getting the input sequences, we use neural event generators consisting of an encoder and a decoder to produce captions. The encoder uses a linear layer and a GRU unit to encode the input sequence~\citep{cho2014properties}. The decoder applies Softmax on the embedding of the input and the hidden state to produce the attention weights. It then uses GRU and a linear layer to produce an English caption of specific objects in the video. Single-object and pairwise captioning models share the same architecture but are trained independently. The hidden dimension of both the encoder and the decoder is 256 for single-object models and 128 for pairwise models. The dropout rate for the decoder is set to 0.3 for single-object models and 0.1 for pairwise models. All models are trained with a learning rate of 0.001. For the grammar check module in the post-processor, we drop the sentences with two consecutively repeated words. We also exclude the sentences that miss verbs or verb arguments, such as sentences ending with words ``from,'' ``to,'' ``at,'' ``is,'' \etc.

\subsection{Baseline Implementation}
\paragraph{Language-Only models.} For the language-only models, we use a LSTM~\citep{hochreiter1997long} with GloVe~\citep{pennington2014glove} word embedding. The hidden dimension of 512, and the dropout rate is 0.2. We use the Adam optimizer~\citep{kingma2014adam} with a learning rate is $4\times 10^{-4}$ and a weight decay of $10^{-5}$. The batch size is 4. Following the data splits in CLEVRER, we split 20\% of the training pairs as the validation set and choose the model with the highest validation accuracy.

\paragraph{CNN+LSTM.} For the CNN+LSTM models, we use a pre-trained ResNet-50 to extract 2,048-dimensional features from the video frames \cite{he2016deep}. We uniformly sample 25 frames for each video as input. The word embedding for questions is initialized by the GloVe~\citep{pennington2014glove} word embedding. Both LSTMs (the question encoder and the video sequence encoder) have 1 layer with a hidden dimension of 512. We apply a dropout rate of 0.2 on the input layer and 0.5 on the hidden layers. We use the Adam~\citep{kingma2014adam} optimizer with a weight decay of $5\times 10^{-4}$. The learning rate is $10^{-5}$ for training from scratch and $10^{-3}$ for finetuning. The batch size is 128 for both trained-from-scratch and finetuning experiments. We split 20\% of the training pairs as the validation set and choose the model with the highest validation accuracy.

\paragraph{BERT+LSTM} We supplement CNN+BERT models as a model with a stronger text encoder. The CNN is the same as in the CNN+LSTM baseline. We use the pretrained BERT uncased base model from HuggingFace library \cite{devlin2018bert}. The BERT tokenizer is set to max length 32 padding and truncation. During training, We fix the weights of the text encoder. We use the Adam optimizer with a weight decay of $5\times 10^{-4}$, and a learning rate of $10^{-5}$ training from scratch and $10^{-3}$ for finetuning. The batch size is 128. We choose the model with the highest validation accuracy with 20\% of the training set as the validation set.

\paragraph{ALOE.} We implement our model based on the publicly released code~\citep{ding2021github}. Since the public release does not contain training code, we implement the training procedure using the following settings. For object embeddings, we use the pre-trained MONet embeddings released by the authors. For optimization, we use the Adam~\citep{kingma2014adam} optimizer with a weight decay of $10^{-3}$ (we have also benchmarked $10^{-2}$, $10^{-3}$ and $10^{-4}$). We split 5\% of the training pairs as the validation set and choose the model with the highest validation accuracy.

\subsection{Error analysis}
\begin{figure}[!tp]
    \centering\small
    \vspace{-1em}
    \includegraphics[width=\textwidth]{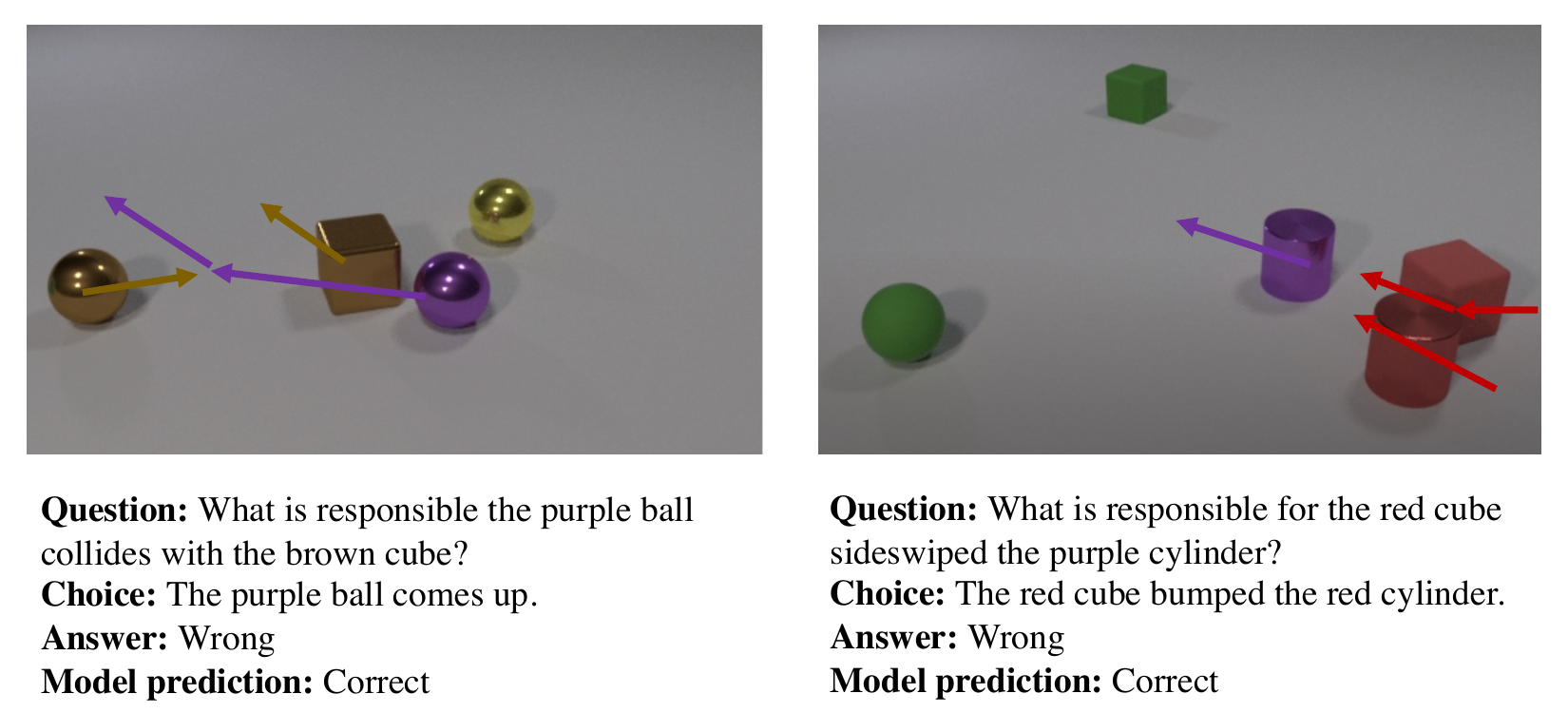}
    \caption{Examples of common prediction errors. The arrows in the image represent the moving direction of objects of interest in the video. \textbf{Left:} failure caused by nuances in human language. While the purple ball is constantly moving upwards coordinate-wise, humans understand the phrase "comes up" as more of the later part of the trajectory (after the purple  ball collides with the brown ball). Therefore, machines cannot give a correct prediction. \textbf{Right:} failure in bridging the domain shift. Humans may consider the change in the trajectory to be minor and appears not to be a deciding factor of the outcome event, but the model predicts it as a cause following similar heuristics in CLEVRER.}
    \label{fig:example_error}
\end{figure} We summarize the common failures of the models: for pretrained-only models, the common error comes from the failure to incorporate more diverse language and events. For example, as shown in Table \ref{tab:parse_error}, the program parser of NS-DR and VRDP fails to generate proper programs for descriptions in \dataset. The deficiency of language understanding often leads to wrong predictions.

\begin{table}[tp]
    \centering\small
    \setlength{\tabcolsep}{10pt}
    \vspace{4pt}
    \begin{tabular}{p{4cm}p{8.7cm}}
    \toprule
       Event & Parsed program \\ \midrule
       The purple sphere slows down from the right. & ["events", "objects", "purple", "filter\_color", "sphere", "filter\_shape", "unique", "filter\_collision", "objects", "unique", "filter\_color", "sphere", "filter\_shape", "unique", "filter\_collision", "unique"]  \\
        The red ball comes to a stop. &  ["events", "objects", "red", "filter\_color", "unique", "filter\_collision", "objects", "red", "filter\_color", "unique", "filter\_collision", "unique"] \\
        The yellow cube comes from the right side at a fast speed. & ["events", "objects", "yellow", "filter\_color", "cube", "filter\_shape", "unique", "filter\_out", "unique"]\\
        \bottomrule
    \end{tabular}
    \vspace{4pt}
    \caption{Examples of errors produced by program parser. In the first row, the model cannot identify the event "slow down from the right" and gives incorrect parsing to find another object involved in a collision ("filter\_collision"). In the second row, the model cannot represent the event "come to a stop" due to the expansion in vocabulary and gives an incorrect output ("filter\_collision"). In the third row, the model mistakenly represents the enter event as the exit event ("filter\_out") because the description is more complicated in \dataset. We follow the notation of programs as in NSDR and VRDP.}
    \label{tab:parse_error}
    \vspace{-2em}
\end{table}

For training from scratch models, one possible reason for the test errors is the nuances in human language. Specifically, models do not only need to identify the objects being referred to but also their physical properties: the cause of "the red cube slows down" can be hard to identify because speed does not appear to be as explicit as other properties, such as colors and shapes. As shown in our comparison between human judgement and heuristics-based causal judgements, the nuance in language can influence human judgments, posing difficulties for machines to ground the events and simulate the reasoning process. For instance, the left figure in \fig{fig:example_error} illustrates the nuances in language resulting in a discrepancy between human judgment and prediction. Moreover, for large models such as ALOE, learning to simulate human reasoning processes from scratch based on very little data can be difficult, especially with a limited training size.

For finetuned models, we have not seen significant improvement brought by the pretraining phase. This is primarily because of the domain gap between human judgement and heuristics-based labeling. Specifically, our human experiments have shown that p(Human | CLEVRER-Heuristic) = 0.62. That is, only 62\% of the event pairs that have been labeled as causal in CLEVRER, are labeled as causal by human annotators. The right figure in \fig{fig:example_error} gives an example of the error caused by the domain shift. Future work may consider other ways of pretraining, such as pretraining on event recognition, which may be more transferable, and pretraining with other types of heuristics.

\clearpage

\section{Labeling Interface}
\label{app:interface}
\begin{figure}[!ht]
    \centering\small
    \includegraphics[width=0.8\textwidth]{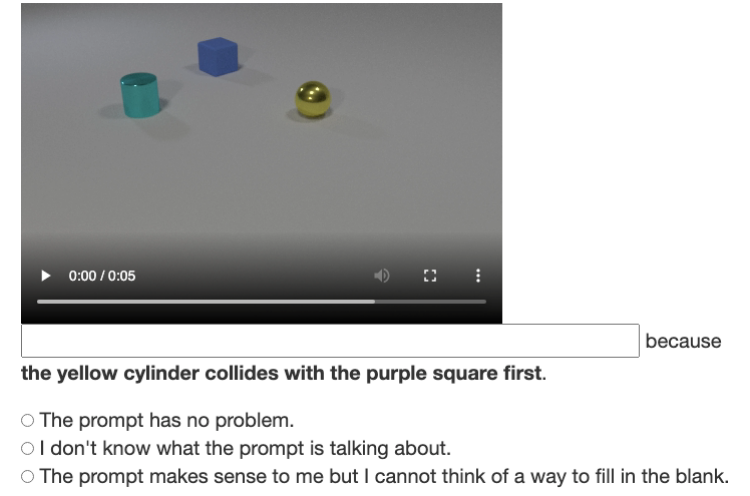}
    \vspace{0em}
    \caption{Example of a causal cloze trial. The participants are asked to fill in the blank after watching a video. They can also select the checkboxes if they do not understand the prompt.}
    \label{fig:cloze_interface}
    \vspace{0em}
\end{figure}

\begin{figure}[!tp]
    \centering\small
    \includegraphics[width=\textwidth]{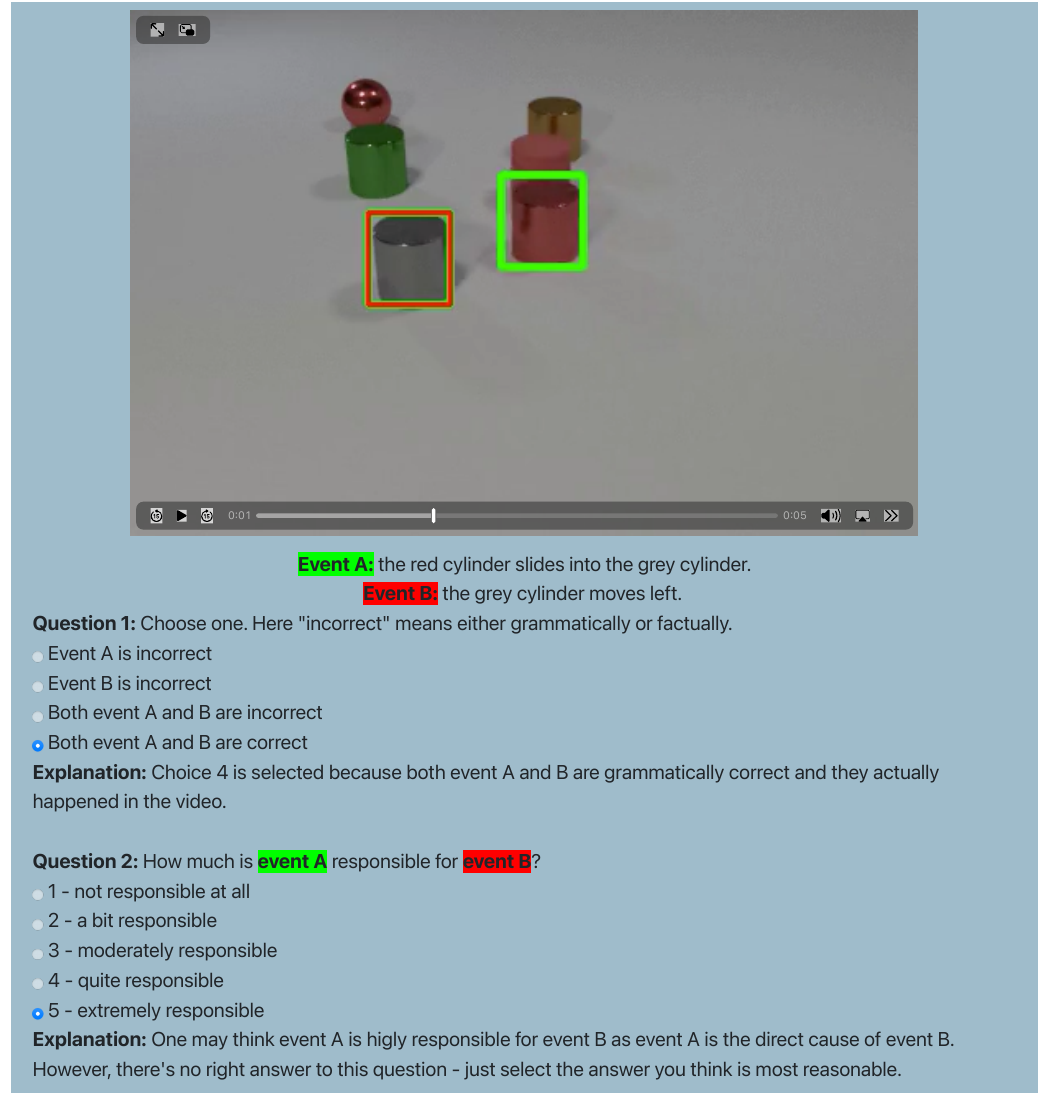}
    \vspace{-1em}
    \caption{Example of a CEG trial. The participants are first asked to select if the event descriptions are correct. If both correct, they are asked to label the level of causal relations between the descriptions. For each event pair, we provide bounding boxes for objects involved in the events for better annotation efficiency.}
    \label{fig:ceg_interface}
    \vspace{0.5em}
\end{figure}

We develop labeling interfaces based on boto3 with Amazon MTurk python API. We include example trials of both the causal cloze tasks and CEG annotation tasks in \fig{fig:cloze_interface} and \fig{fig:ceg_interface}, respectively. The full instruction texts are provided on the labeling page of our project's website. The estimated hourly pay to the Mechanical Turk participants is about \$6.1, and the total amount spent on participant compensation is about \$3500. Specifically, the cloze tests and part of the pairwise causal relationship annotations were completed by users from the U.S., and the pay was \$7.7/hour (above the federal minimum wage). At a later stage of our project, we were unfortunately constrained by the budget available to us and opened the tasks to workers outside the U.S. Thus, overall, our average hourly wage is \$6.1. Our goal has always been to commit to best practices and offer fair pay to users whenever possible, and we will continue to do so in the future. 

In causal cloze tasks, the participants are asked to write an event description given a cause or outcome event, as shown in \fig{fig:cloze_interface}. We specify the expectation of responses (such as using complete sentences, avoiding ambiguous third-person pronouns, \etc) in the instruction. We design a small comprehension quiz with 7 multiple-choice questions and 2 chances to submit to ensure the participants understand the instructions correctly. 

In the CEG annotation tasks, the participants are asked to label the correctness and causal relation of two event descriptions as shown in \fig{fig:ceg_interface}. We give 4 examples with detailed explanations to help them understand the rationale of the task. We also give an illustration of object colors and shapes for reference. Bounding boxes are added to the videos to accelerate the process of locating objects of interest.

\clearpage

\section{Dataset Release}
\label{app:sheet}
Our dataset is under CC0 License. We provide a documentation using data statements for NLP in \cite{10.1162/tacl_a_00041}. 

\paragraph{Short form data statement} \dataset is a large-scale video reasoning dataset of human-annotated physical event descriptions and their causal relations. It contains machine-generated texts based on crowdsourcing data in US English (en-US). The language quality and causal structure annotations are obtained by watching videos, reading texts, and entering responses on MTurk.

The following is the long-form data statement of \dataset:
\paragraph{Curation Nationale}
\dataset contains descriptions and causal relations of physical events such as an object entering the scene and two objects colliding with each other. The goals in selecting texts were to ensure the interpretability and correctness of the descriptions and to provide a variety of free-form captioning of physical events in videos. We first collected human-written event descriptions by causal cloze tasks, then used machine learning models to generate more natural language descriptions based on the curated data. We post-processed the data by grammar checking, object existence checking, and verb re-balancing. Finally, we obtained human annotations on the texts through crowdsourcing: if the labelers annotated the texts as interpretable and correct, we asked them to provide a pairwise graded causal judgment of the events.

\paragraph{Language Variety}
The event description data for causal cloze tasks were collected on MTurk. Information about which varieties of English are represented is not available, but at least \dataset includes US (en-US) mainstream English.

\paragraph{Speaker Demographic}
We used a cascaded generator composed of a rule-based event detector and a neural pairwise generator to generate texts. When curating training data in causal cloze tasks, we restricted the location of these MTurkers to be in the US. It is expected that most speakers use English as their native language.  Estimated demographics of MTurkers may refer to \cite{ross2010crowdworkers}.
\paragraph{Annotator Demographic}

We hire MTurkers with approved HITs of 1000 or higher. We expect the MTurkers to be the general public who are familiar with the basic crowdsourcing process. When collecting data, we release the tasks in batches, where each HIT contains 30 QA pairs, mostly coming from one or two videos. We perform a quality check to ensure annotators have sufficient knowledge of the English language. We also answer their questions about the annotation process by email. It is expected that most speakers use English as their native language. Estimated demographics of annotators may refer to \cite{ross2010crowdworkers}. 

\paragraph{Speech Situation}
The intended audience of the texts is the general public. The texts are all in written form. MTurkers are expected to read the text and watch the video when doing causal cloze and causal labeling tasks. The video is about 5 seconds, which can be played as many times as one wishes. 
\paragraph{Text Characteristics} 
The texts are plain English descriptions of a physical event in a video. A sentence usually contains one or more physical object(s) (i.e. sphere, cylinder, or cube) and the related movements or interactions presented in the video. Ideally, the generated event descriptions can maintain the vocabulary and structural characteristics similar to the training data from causal cloze tasks. The detailed statistics of the text are shown in the Dataset Statistics section.
\paragraph{Recording Quality} N/A
\paragraph{Other} N/A
\paragraph{Provenance Appendix}  The videos of \dataset are from the CLEVRER dataset~\citep{yi2019clevrer}.
\subsection{Intended Use}
CLEVRER-Humans can be used as a benchmark in physical scene understanding and causal reasoning. It evaluates machines' ability to understand and analyze physical interventions in a restricted setting. Machines are provided with a short video and expected to answer questions regarding the causes of events in the video.
\subsection{Maintenance Plan}
We will host our dataset permanently on our project's website. Users are granted access to the dataset through links on the website. We provide versioning of the dataset and archive backup regularly.

\subsection{Quality Check}
Quality checks over CEG node correctness are performed by majority voting. Since we have allocated the annotation of each video to 3 annotators, and they will see overlapping events and annotate their correctness. Checks for edge correctness are performed by including additional “quality checking” questions. Specifically, each annotator will see 3 videos and 10 questions for each video. 1 of the video will be from a small and manually-curated dataset by authors, containing 30 videos. The entire answer set will be accepted if and only if the annotators answer those quality-checking questions correctly (more specifically, have a small divergence from our answer).
 
\end{document}